\pdfoutput=1   
\documentclass[letterpaper]{article} 
\usepackage[preprint]{aaai2027}   
\usepackage[hyphens]{url}     
\usepackage{graphicx}         
\usepackage{natbib}           
\usepackage{caption}          
\usepackage{amsmath,amssymb,amsfonts}
\usepackage{booktabs}         
\usepackage{multirow}


\newcommand{\NValHM}{2000}  
\newcommand{\NScenesHM}{20}  
\newcommand{\NCatsHM}{six}  
\newcommand{\NValMP}{2195}  
\newcommand{\NScenesMP}{11}  
\newcommand{\NCatsMP}{21}  
\newcommand{\SuccDist}{1.0}  
\newcommand{\MaxSteps}{500}  

\newcommand{\OursMPSR}{47.29}  
\newcommand{\OursMPSPL}{18.01}  
\newcommand{\OursHMSR}{68.75}  
\newcommand{\OursHMSPL}{33.97}  
\newcommand{\NeverHMSR}{65.35}  
\newcommand{\NeverHMSPL}{32.78}  
\newcommand{\AlwaysHMSR}{64.80}  
\newcommand{\AlwaysHMSPL}{31.36}  
\newcommand{\OneShotHMSR}{65.05}  
\newcommand{\OneShotHMSPL}{31.80}  

\newcommand{\NoMemHMSR}{65.40}  
\newcommand{\NoMemHMSPL}{32.49}  

\newcommand{\NoRescanHMSR}{68.50}  
\newcommand{\NoRescanHMSPL}{33.92}  

\newcommand{\DeltaSRHM}{3.40}  
\newcommand{\DeltaSRHMCILo}{1.70}  
\newcommand{\DeltaSRHMCIHi}{5.05}  
\newcommand{\DeltaSROneShot}{3.70}  
\newcommand{\DeltaSROneShotCILo}{2.00}  
\newcommand{\DeltaSROneShotCIHi}{5.45}  
\newcommand{\DeltaSRNoMem}{3.35}  
\newcommand{\DeltaSRNoMemCILo}{1.75}  
\newcommand{\DeltaSRNoMemCIHi}{5.00}  
\newcommand{\DeltaSRAlways}{3.95}  
\newcommand{\DeltaSRAlwaysCILo}{2.20}  
\newcommand{\DeltaSRAlwaysCIHi}{5.65}  

\newcommand{\McNemarChiHM}{15.06}  
\newcommand{\McNemarPHM}{<0.001}  

\newcommand{\OursHMCalls}{5.00}  
\newcommand{\OursHMImg}{3.04}  
\newcommand{\OursHMThink}{17.2}  
\newcommand{\NeverHMCalls}{1.49}  
\newcommand{\NeverHMImg}{1.49}  
\newcommand{\NeverHMThink}{5.7}  
\newcommand{\AlwaysHMCalls}{8.87}  
\newcommand{\AlwaysHMImg}{4.88}  
\newcommand{\AlwaysHMThink}{24.3}  
\newcommand{\OneShotHMCalls}{3.75}  
\newcommand{\OneShotHMImg}{3.75}  
\newcommand{\OneShotHMThink}{17.4}  
\newcommand{\NoMemHMCalls}{4.99}  
\newcommand{\NoMemHMImg}{3.02}  
\newcommand{\NoMemHMThink}{18.5}  
\newcommand{\NoRescanHMCalls}{5.00}  
\newcommand{\NoRescanHMImg}{3.04}  
\newcommand{\NoRescanHMThink}{17.2}  

\newcommand{\OursStepsPerEp}{205}  

\newcommand{\RescanTrigRate}{6.1}  
\newcommand{\RescanYieldRate}{41.4}  
\newcommand{\VerifyVetoRate}{16.7}  

\newcommand{\ZSONHMSR}{25.5}  
\newcommand{\ZSONHMSPL}{12.6}  
\newcommand{\ZSONMPSR}{15.3}  
\newcommand{\ZSONMPSPL}{4.8}  
\newcommand{\ESCHMSR}{39.2}  
\newcommand{\ESCHMSPL}{22.3}  
\newcommand{\ESCMPSR}{28.7}  
\newcommand{\ESCMPSPL}{14.2}  
\newcommand{\LThreeMVNHMSR}{50.4}  
\newcommand{\LThreeMVNHMSPL}{23.1}  
\newcommand{\LThreeMVNMPSR}{34.9}  
\newcommand{\LThreeMVNMPSPL}{14.5}  
\newcommand{\VLFMHMSR}{52.5}  
\newcommand{\VLFMHMSPL}{30.4}  
\newcommand{\VLFMMPSR}{36.4}  
\newcommand{\VLFMMPSPL}{17.5}  
\newcommand{\SGNavHMSR}{54.0}  
\newcommand{\SGNavHMSPL}{24.9}  
\newcommand{\SGNavMPSR}{40.2}  
\newcommand{\SGNavMPSPL}{16.0}  
\newcommand{\ASCENTHMSR}{65.4}  
\newcommand{\ASCENTHMSPL}{33.5}  
\newcommand{\ASCENTMPSR}{44.5}  
\newcommand{\ASCENTMPSPL}{15.5}  

\newcommand{\GainDelibFP}{1.35}  
\newcommand{\GainDelibNF}{1.25}  
\newcommand{\GainDelibFN}{1.00}  
\newcommand{\GainDelibTotal}{3.40}  
\newcommand{\GainDelibPos}{3.60}  
\newcommand{\GainMemNF}{2.55}  
\newcommand{\GainMemTotal}{3.35}  

\newcommand{\ind}[1]{\mathbf{1}[#1]}

\title{Hierarchical Fast--Slow ReAct Agent for\\Zero-Shot Object-Goal Navigation}

\author{
    Zhaochen Lan\textsuperscript{\rm 1},
    Zhi Yang\textsuperscript{\rm 2},
    Yuxiang Fu\textsuperscript{\rm 1},
    Mengxiang Lin\textsuperscript{\rm 1}\corresponding
}
\affiliations{
    \textsuperscript{\rm 1}School of Mechanical Engineering and Automation,
    Beihang University, Beijing, China\\
    \textsuperscript{\rm 2}School of Automation,
    Beijing Institute of Technology, Beijing, China\\
    lanzhaochen@buaa.edu.cn, linmx@buaa.edu.cn
}

\begin{document}

\maketitle

\begin{abstract}
Zero-shot object-goal navigation (ZSON) requires a robot to find a named
object category in a building it has never entered. The prevailing approach
scores frontiers with a vision--language \emph{value map}: every decision is
another argmax over the map as it currently stands, and the evidence behind
that score is discarded the moment it is taken. Systems that place a large
vision--language model inside the perception--action loop typically query it on a fixed
schedule from the current view alone; a room the robot walked through minutes
earlier is never reconsidered, and a failed call has no defined fallback.
We turn what the robot has already seen into the object of deliberation. Our
hierarchical fast--slow agent leaves the value-map controller running at every
step and writes a \emph{coordinate-anchored memory} as it moves: a semantic
grid of room types and confirmed object instances, together with a bounded
store of pose-tagged keyframes. A VLM screens each candidate detection before
it is written. A deliberative layer reads this memory in a bounded
reason--retrieve--act loop. It wakes on structural events the reactive
layer computes, reasons first over text, and recalls a first-person view
only for candidates that text alone cannot separate. Per-invocation and
per-run caps bound its calls, a call-free first tier resolves the most
frequent stall, and any failure returns control to the reactive controller.
Our system reaches \OursHMSR\% SR on HM3D v1
\texttt{val} and \OursMPSR\% on MP3D \texttt{val}, the highest success rate
among the zero-shot methods compared here. Choosing among far frontiers by
argmax instead of deliberating costs \DeltaSRHM{} SR points in a paired
comparison over all \NValHM{} HM3D episodes (95\% CI [\DeltaSRHMCILo,
\DeltaSRHMCIHi]); deliberating over every frontier does not recover them.

\end{abstract}


\section{Introduction}
\label{sec:intro}

Zero-shot object-goal navigation (ZSON) places a robot at a random pose in a
previously unseen indoor environment and asks it to find an instance of a
named object category---\emph{bed}, \emph{toilet}, \emph{plant}---without any
category-specific training~\citep{objectnav,zson}. A prominent family of
solutions explores reactively, scoring each frontier between explored and
unknown space with a vision--language \emph{value map} that measures how
much the view toward it resembles the goal~\citep{vlfm,esc,l3mvn,ascent}.

What such a controller lacks is a way to act when it has no action it
believes in: clutter defeats the local planner, the only frontiers left lie
far away, the frontier set runs out because an opening was missed, or the goal
sits on another floor. These are \emph{structural impasses}---cheap for the
reactive layer to detect, and beyond what another argmax over its current
value map can resolve.
\begin{figure*}[t]
\centering
\includegraphics[width=0.98\textwidth]{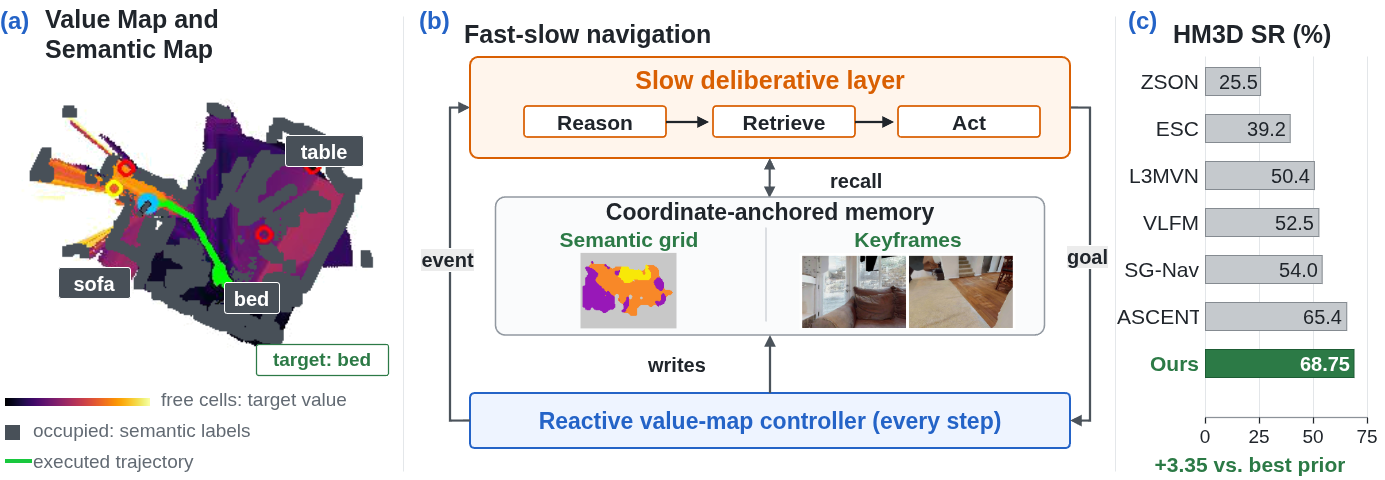}
\caption{\textbf{Hierarchical fast--slow navigation at a glance.}
\emph{(a)}~One HM3D run with \emph{bed} as the goal: the reactive value-map
controller scores free cells by how much the view toward them resembles the
goal (colour bar), labels occupied cells as it recognises them, and executes
the green trajectory. \emph{(b)}~That controller runs at every step and writes
a coordinate-anchored memory---a semantic grid and a store of pose-tagged
keyframes. A bounded reason--retrieve--act loop wakes on the structural events
the controller reports, recalls what it needs, and returns a
new goal. \emph{(c)}~The full system reaches \OursHMSR\% SR on HM3D v1
\texttt{val}, the highest among the zero-shot methods we compare.}
\label{fig:teaser}
\end{figure*}
Three of the four turn on an earlier observation the
controller no longer keeps. Systems that bring large language and vision--language models into
navigation~\citep{cognav,sgnav} call the model on a schedule fixed before the
robot moves, and hand it evidence fixed with that schedule: a room the robot
walked through minutes earlier cannot be re-examined, since a summary of it may
survive but the view itself does not. And how that
consultation is governed---when to call, how much evidence to gather, where
control goes when the call fails---is left unspecified.

Our agent is hierarchical and fast--slow (Fig.~\ref{fig:teaser}): a reactive value-map controller runs at
every step and, as it explores, writes a \emph{coordinate-anchored memory}---a semantic grid of room types and confirmed object instances, plus a bounded store of pose-tagged keyframes. A goal detection---the only evidence that can authorize stopping---passes a verification gate that vetoes it \emph{before} fusion makes it permanent.
A bounded
reason--retrieve--act loop~\citep{react} reads that memory: it wakes on the structural events the reactive layer computes, reasons first over text, and pulls up a
remembered first-person view only for candidates that text alone cannot
distinguish.
Two-level budgets cap every loop, and every failure path returns control to
the reactive layer, which stays available until it reaches its own terminal
condition.

On HM3D v1 and MP3D \texttt{val}~\citep{hm3d,mp3d} the full system leads
every zero-shot baseline of Table~\ref{tab:main}. Choosing among far frontiers
by argmax instead of deliberating costs \DeltaSRHM{} points of SR, while
deliberating over every frontier recovers none of that loss.

Our contributions are:
\begin{enumerate}
\item \textbf{A hierarchical fast--slow agent for ZSON.} A reactive value-map
controller keeps the robot moving at every step, while a deliberative ReAct
layer sits off that per-step path and intervenes only at structural impasses:
deliberation redirects the robot; it is never what drives it.
\item \textbf{A coordinate-anchored memory with an abstract and a literal
tier.} A semantic grid accumulates room types and object instances over the
whole run; a bounded store keeps the pose-tagged views themselves; one
coordinate frame indexes both, so the deliberative layer reasons over text
first and opens the view behind a claim only when text cannot decide. Since
neither tier can undo a fused observation, each admits evidence by what that
evidence can cause: an instance becomes retrievable after two sightings, and a
goal detection must pass a per-detection verification gate before fusion.
\item \textbf{An invocation policy that makes deliberation bounded and
fail-open.} When to call, how much evidence one call may gather, and where
control goes when it fails are parts of the method rather than deployment
details: the trigger is a structural impasse the reactive layer already
computes, a call-free first tier answers the most frequent stall,
per-invocation and per-run budgets cap the rest, and every failure path
returns control to the reactive controller.
\end{enumerate}

\section{Related Work}
\label{sec:related}

\paragraph{Zero-shot object-goal navigation}
The ObjectNav task definition and evaluation protocol were consolidated
by \citet{objectnav}, with progress measured in the Habitat
simulator~\citep{habitat} on the HM3D~\citep{hm3d} and Matterport3D
(MP3D)~\citep{mp3d} scene datasets, where frontier-based
exploration~\citep{yamauchi} and learned semantic-map policies~\citep{semexp}
set the early benchmark. The zero-shot line removes
category-specific training: CoW~\citep{cow} drives frontier exploration with an
open-vocabulary detector; ZSON~\citep{zson} transfers image-goal training through multimodal
goal embeddings; ESC~\citep{esc} and L3MVN~\citep{l3mvn} inject language-model
commonsense into frontier selection; VLFM~\citep{vlfm} scores frontiers
directly with a vision--language value map; and ASCENT~\citep{ascent} extends
the value-map recipe with floor-aware, coarse-to-fine exploration. All of
these controllers are reactive: each decision consumes the current map and
observation through a fixed scoring rule, and when that rule reaches a
structural impasse there is no deliberative recourse. Nor is there a check between
detecting a goal candidate and writing it into the map: VLFM and ASCENT fuse
what the detector reports.

\paragraph{LLM/VLM-guided navigation and scene-graph reasoning}
A second line consults a large model during navigation. CogNav~\citep{cognav}
models the navigation process as transitions among cognitive states
orchestrated by an LLM; SG-Nav~\citep{sgnav} incrementally builds an online
hierarchical 3D scene graph and prompts an LLM with it for zero-shot object
navigation. In both, \emph{when} the model is called is settled before the
robot moves---every step, or on a prespecified sequence of states---and every
call is handed the same kind of evidence. What the robot saw earlier reaches the
model only as whatever its representation chose to keep---SG-Nav's graph nodes,
or, in image-goal navigation, the trajectory fragments MemoNav~\citep{memonav}
judges informative---never as the view itself. Spatial memories built for language queries---VLMaps~\citep{vlmaps} fusing
vision--language features into a 3D map, ConceptGraphs~\citep{conceptgraphs}
into an open-vocabulary 3D scene graph---keep descriptors of a place rather
than the views of it, and neither decides when a model should be
consulted. Our controller instead picks its call times at run time, from four
structural impasses the reactive layer detects as it runs
(Section~\ref{sec:trigger}).

\paragraph{ReAct agents}
A reason--act agent interleaves reasoning, tool calls, and the observations
those calls return, so the model itself decides at run time what to examine
next~\citep{react}. Agents of this shape now close real GitHub
issues~\citep{swebench,sweagent} and run chemistry experiments on laboratory
hardware~\citep{coscientist,chemcrow}.
Embodied work has adopted the loop by handing it the run: ORION~\citep{orion}
emits thought--action pairs over perception and memory tools to drive a Habitat
robot through dialogue-driven personalized search; TANGO~\citep{tango} has an
LLM compose navigation and perception primitives into a program written once,
before the robot moves; and CogNav~\citep{cognav} re-queries the model at every
decision to advance a cognitive state machine. We put the loop beneath the
reactive controller instead of in front of it: the controller drives every step
unaided, the loop wakes only at a structural impasse, and the model then chooses
for itself which part of the anchored memory to read.
Section~\ref{sec:ablations} contrasts the two placements directly.

\section{Method}
\label{sec:method}

We address zero-shot object-goal navigation~\citep{objectnav}: a robot placed
at a random pose in an unseen, multi-room, possibly multi-floor building must
find an instance of a category $g$. At each step it receives an egocentric RGB-D
observation and an odometry pose $\mathbf{p}_t=(x_t,y_t,\psi_t)$, and emits a
discrete action.

\subsection{Architecture Overview}
\label{sec:overview}

\begin{figure*}[t]
\centering
\includegraphics[width=1.0\textwidth]{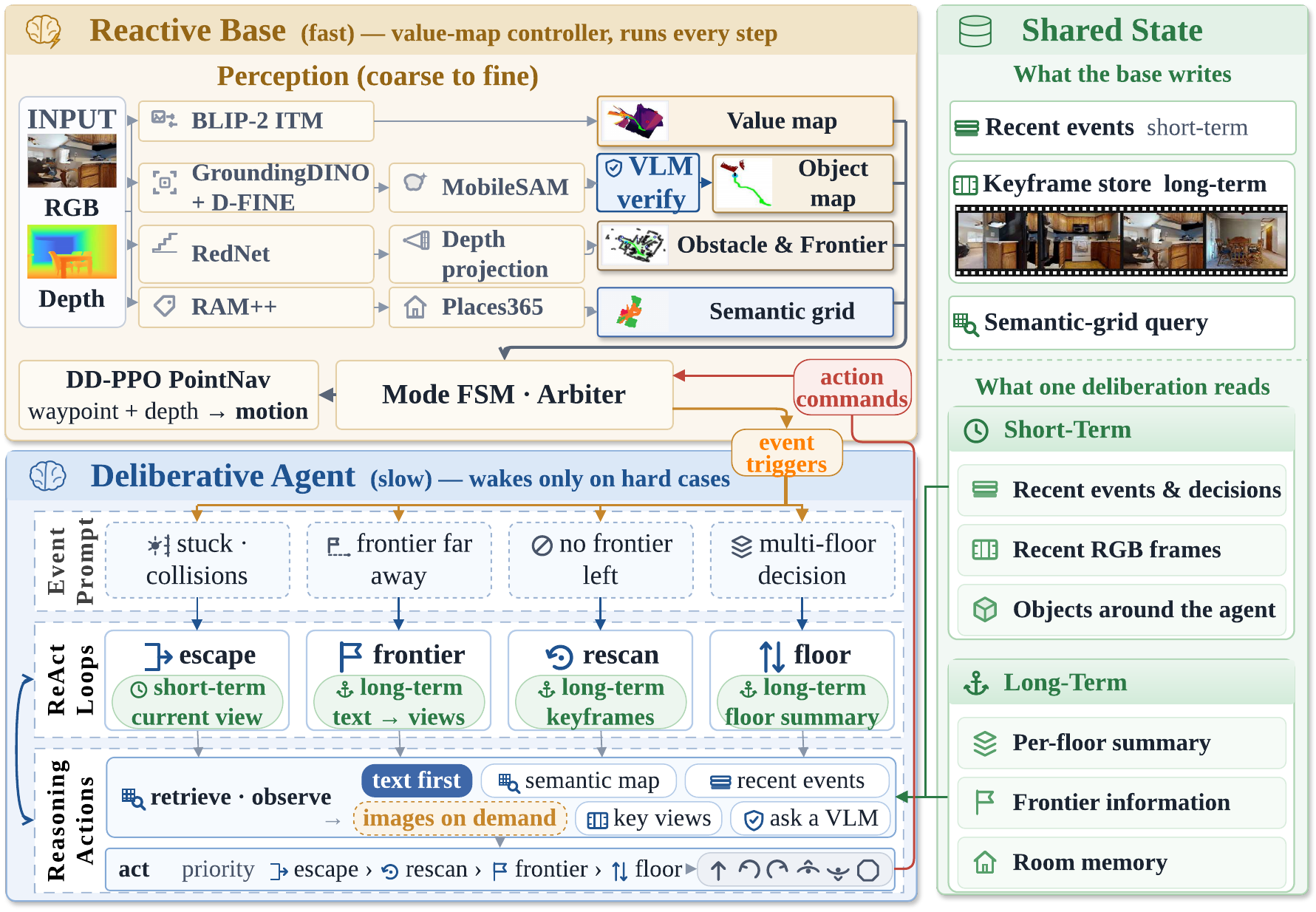}
\caption{System architecture. The always-on \emph{Reactive Base} (top, fast)
perceives, maps, and moves at every step and writes the shared state
(right)---short-term event context plus the coordinate-anchored memory itself,
the semantic grid and the keyframe store. An
event-triggered \emph{deliberative agent} (bottom, slow) wakes only at a
structural impasse, reads that memory in a bounded ReAct loop---annotated with
the horizon it queries---and returns typed commands under strict budgets. Any
deliberative failure degrades back to the Reactive Base, which executes every
motion primitive.}
\label{fig:architecture}
\end{figure*}

The agent has two controllers sharing spatial state
(Figure~\ref{fig:architecture}). The \emph{reactive controller} runs
continuously, handling perception, mapping, and short-horizon motion; it
builds on ASCENT~\citep{ascent}, a value-map controller in the
VLFM~\citep{vlfm} family, augmented with an online per-detection VLM
verification gate and a triggered slow-thinking brain. A \emph{deliberative
controller} runs only when a triggering event fires: it then executes a
bounded reason--retrieve--act loop~\citep{react} and returns a typed command
the reactive controller executes. The two couple at the data level: the
reactive controller stores what it observes, and the deliberative controller
turns selected entries into typed navigation targets.

Three independent switches set the deliberative layer's behaviour: \emph{when}
it fires, \emph{how} its evidence budget is spent, and \emph{which}
memory-dependent capabilities are enabled. Section~\ref{sec:ablations} turns
each off in turn.

\subsection{Reactive Controller}
\label{sec:reactive}
The reactive controller maintains, per floor, four $2$D maps built by
projecting depth into the odometry frame. An \emph{occupancy map} supports
collision-free motion and frontier extraction. A \emph{value map} stores in
every free cell $c$ the target-conditioned score
$V(c)=\operatorname{sim}(\phi_{\text{img}}(I_c),\phi_{\text{txt}}(g))$, with
$\phi_{\text{img}},\phi_{\text{txt}}$ the encoders of a pretrained
image--text matching model~\citep{blip2} and $I_c$ the view associated with
$c$. A \emph{semantic grid} (Section~\ref{sec:memory}) records
\emph{what} was seen and \emph{where}; the reactive pipeline populates it from
quantities it already computes---no extra model call---and the deliberative
controller alone consumes it. An \emph{object map} accumulates the goal
detections that pass the gate below---the map that can authorize stopping.
Frontier points on the boundary with the unknown, read off the occupancy map,
give the candidate exploration targets.

Goal candidates come from a detection ensemble~\citep{dfine,gdino,ram} and
a lightweight segmenter~\citep{mobilesam}, room labels from a scene
classifier~\citep{places365} and pixel semantics from an RGB-D
network~\citep{rednet}; a pretrained point-goal policy~\citep{ddppo} executes
motion toward a chosen target, and inter-floor transitions are handled by a
stair finite-state machine. Every candidate detection then passes an
\emph{online per-detection verification gate} before it is fused into the
object map. The gate shows a VLM the current frame cropped to the candidate
box together with the proposed category, and returns an accept/reject decision;
rejected candidates are discarded and may be proposed again from a later
frame. Because a fused cloud merges detections by
category, no individual detection can be removed once fused; verifying first
keeps rejected candidates out of the map that authorizes stopping, whereas a
later check could only mark them as suspect.

\subsection{Coordinate-Anchored Memory}
\label{sec:memory}

\begin{figure*}[t]
\centering
\includegraphics[width=1.0\textwidth]{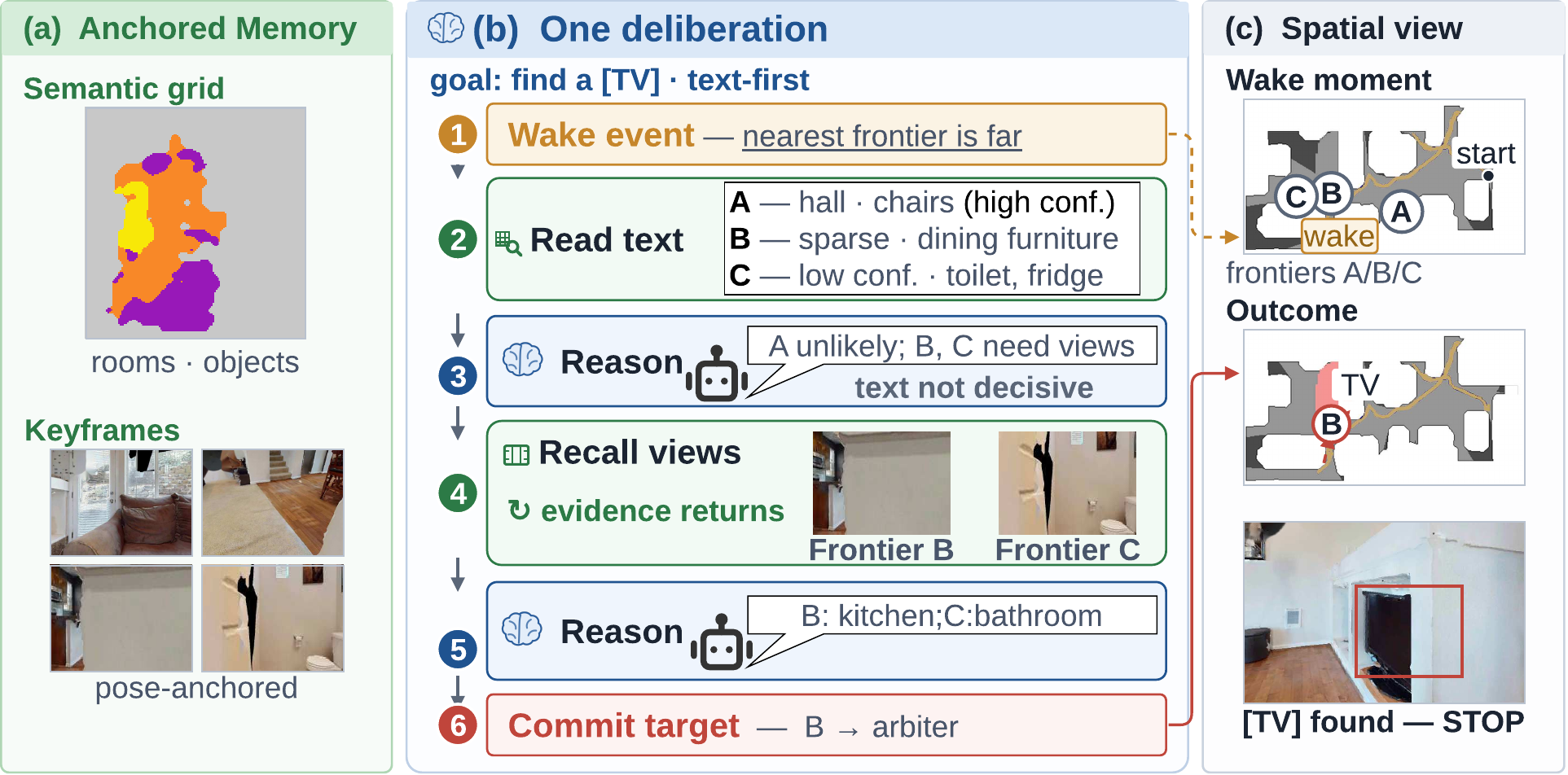}
\caption{One deliberation, end to end, on one HM3D evaluation run (goal: TV,
success). \emph{(a)}~Anchored Memory---a semantic grid, tinted by per-cell
room type (here hall in orange, kitchen in purple, dining room in yellow,
over light grey for space not yet observed), and
pose-anchored keyframes---is written by the Reactive Base as it explores and
reset at the start of every run. \emph{(b)}~The nearest frontier lies far, so
the bounded reason--retrieve--act loop wakes and reads textual memory first.
When that round cannot separate frontiers B and C, it retrieves one stored view
toward each---remembered, not re-visited---submits those images in a second
round, and commits a single typed target. \emph{(c)}~The wake moment in space: trajectory (tan),
wake pose, and frontiers A/B/C; B opens onto a kitchen--living area, and
driving there reveals the TV (red box).}
\label{fig:memory}
\end{figure*}

The reactive controller stores room labels, confirmed objects, and pose-tagged
keyframes in two coordinate-indexed structures, shown architecturally in
Figure~\ref{fig:architecture} and on an example in Figure~\ref{fig:memory}(a).

\paragraph{Semantic grid}
We maintain a coarse grid whose cell $\Pi(\mathbf{x}) = \lfloor
\Pi_{\text{px}}(\mathbf{x})/8 \rfloor$ aggregates an $8{\times}8$ block of
occupancy pixels, reusing the occupancy map's own world$\rightarrow$pixel
transform $\Pi_{\text{px}}$ to guarantee consistency. Each cell stores (i) a
room-type vote histogram, accumulated rather than overwritten so that
transient misclassifications average out; (ii) a bounded ring of
\emph{keyframe indices} whose view frusta covered the cell; and (iii) a
sparse table of \emph{object instances} (class, back-projected world position,
observation count, supporting keyframes). These are context objects---categories
other than the goal, which the reactive controller tracks separately in its
object map---paired with the views that support them, so that deliberation can
retrieve them by location. We call an instance
\emph{confirmed}, and make it retrievable, once it has been observed at least
twice, which suppresses spurious one-shot detections. At $0.4$\,m cells, retrieval keyed by world
coordinates tolerates pose jitter; because every cell and keyframe also
carries its floor index, a query can reach floors the robot has left.

\paragraph{Keyframe store}
In parallel we keep a bounded store of low-resolution keyframe thumbnails,
sampled on translation/rotation thresholds, each tagged with its pose,
floor, and step index and back-filled with the room label and visible
objects. Eviction enforces a \emph{per-floor} capacity, so a newly entered
floor cannot evict the memory of earlier ones.

\paragraph{Retrieval interface}
Given a query world position, the memory returns four things: the local room
estimate and its confidence, the nearby confirmed instances, a coverage
statistic (how much of this neighborhood has actually been seen), and a ranked
set of candidate keyframes. Candidates are the cell's view-footprints together
with the supporting views of nearby instances, passed through a field-of-view
gate and a distance band and ranked by
$s(k)=\cos(\Delta\beta_k)-\lambda d_k$ subject to
$|\Delta\beta_k|\le\frac{1}{2}\text{HFOV}$, with $\Delta\beta_k$ the bearing
offset of keyframe $k$ from the query and $d_k$ its distance. Image recall
follows a strict fallback ladder---exact footprint hit, relaxed footprint
search, the
frontier's discovery frame, and finally an empty result telling the caller
to proceed text-only. A retrieval never interrupts control: when no view can be
recalled the loop continues on a step-indexed context and then a floor-level
summary.

\subsection{Event-Triggered Deliberation}
\label{sec:trigger}
Four structural impasses wake the deliberative controller, each detected from
quantities the reactive controller already maintains, and each answered by its
own bounded loop. Every invocation obeys a per-invocation budget---at most
$M{=}3$ reasoning rounds, at most two images in any one call, and a wall-clock
cap---on top of the per-run budget of Section~\ref{sec:budget}.

\paragraph{Contact stalls}
From the platform's cumulative collision count $n_t$ we form a per-step
contact indicator over a sliding window $W$ and call the robot stuck when
$\sum_{\tau=t-W+1}^{t}\ind{n_\tau>n_{\tau-1}}\ge\kappa$, with a rising-edge
latch and a cooldown so one physical stall yields one event; a
no-net-displacement odometry test over a longer window is a secondary signal.
Both are suspended while the controller is already deliberating, so a
deliberate pause is not mistaken for a stall.
Recovery is then tiered, so that the common stall costs no model call at all.
The first tier suppresses the pursued frontier target and forces a replan,
which resolves stalls caused by an unreachable or ill-chosen target. Only if
the robot fails to move away within a fixed displacement window, or stalls
again under the suppression, does the controller escalate: a VLM is queried
with the current view for an escape heading, and the matching short
macro-action is executed. Sufficient net motion ends the loop; otherwise the
query repeats with previously tried directions as explicit negatives, up to
the round budget. Exhaustion disables the offending region and falls through
to the reactive base's own recovery.

\paragraph{Distant frontiers}
Committing to a far frontier costs dozens of steps to undo, and a similarity
heatmap alone offers no way to tell two distant frontiers apart. With
frontier candidates $F=\{\mathbf{f}_i\}$ and robot position $\mathbf{p}$, the
controller deliberates when even the nearest candidate is beyond a distance
gate, $\min_i \lVert \mathbf{f}_i-\mathbf{p}\rVert > d_{\text{far}}$, subject
to the same cooldown and budget checks; near frontiers keep the cheap
reactive path.
The loop acquires \emph{visual evidence only when text is insufficient}
(Figure~\ref{fig:memory}). For each of the nearest far candidates it
assembles a textual description from the semantic memory (room estimate,
nearby objects) and the target's room affinities, read from a fixed
category--room knowledge graph. The first round is
\emph{text only}: the model either commits to a candidate or, if the
descriptions are too ambiguous, requests first-person images for at most two
named candidates. Only then does the controller retrieve the best-available
keyframe per requested candidate by $s(k)$ and issue a second, image-bearing
round, which turns a fixed image budget into adaptive, per-decision
expenditure. The chosen frontier becomes a persistent deliberative
target (Section~\ref{sec:budget}); a parsing failure or exhausted budget
returns control to the reactive value-ranking path.

\paragraph{Frontier exhaustion}
When the frontier set empties before the goal is found, the reactive base has
nothing left to explore and would end the run---exactly when a missed opening
is most consequential and what the robot has already seen is most
informative. Rather than accept that stop, the controller re-examines the
current floor's stored keyframes---remembered observations, not a physical
re-traverse---subsampled uniformly into one indexed montage, and asks in a
single image-bearing call whether any remembered view shows an overlooked
passage (staircase, door, opening) into unexplored space. An identified view
turns its anchored pose into a persistent goal with a validity horizon, and
exploration targets previously suppressed near that pose are re-enabled so
arriving there can regenerate frontiers. If the model declines, the reply is
unparsable, or the budget gate is closed, the run ends exactly as it would
have without rescan. Rescan is capped per floor and draws from the shared
per-run budget.

\paragraph{Floor changes}
When exploration on the current floor stalls and other floors are known, the
evidence that settles the choice sits on floors the robot has left. The
controller sends a purely textual per-floor summary to the VLM service to
decide whether to change floors, again fail-open.

\subsection{Arbitration and Budget Control}
\label{sec:budget}
A single arbiter reconciles reactive motion with deliberative commands under
a fixed priority order: a recovery macro-action preempts everything; the stair
finite-state machine retains control during inter-floor transitions; otherwise
navigation proceeds toward a detected goal or a deliberative target. Targets from the distant-frontier loop or rescan
are \emph{persistent}: the arbiter suppresses cheap reactive preemption
until the target is reached or its validity horizon elapses, so the
controller deliberates once and then commits rather than thrashing.

A \emph{model call} is any invocation of the external VLM service by the
deliberative controller, and \emph{image-bearing} calls are the subset carrying
one or more images; the verification gate and the floor decision call the same
service under caps of their own and are counted separately. Calls are
bounded at two levels: the per-invocation caps above, and a \emph{per-run} cap
of $20$ model calls with a tighter cap of $10$ image-bearing calls, shared
across all deliberative loops. A per-step watchdog degrades
the controller for the rest of the run if any single decision exceeds
$180$\,s, and a circuit-breaker disables model calls after consecutive
failures. Degradation is deterministic rather than best-effort: every failure
path---empty memory, service outage, unparsable reply, exhausted
budget---has the reactive base as its single terminal state, so no
deliberative failure can leave the agent without a controller.
Table~\ref{tab:triggers} in the appendix summarizes modes, triggers, and
budgets.

\section{Experiments}
\label{sec:experiments}

\subsection{Benchmarks and Metrics}
\label{sec:protocol}

\begin{table}[t]

\centering
\small
\setlength{\tabcolsep}{0.8mm}
\begin{tabular}{@{}l cc cc@{}}
\toprule
& \multicolumn{2}{c}{HM3D v1} & \multicolumn{2}{c}{MP3D} \\
\cmidrule(lr){2-3} \cmidrule(l){4-5}
Method & SR\,$\uparrow$ & SPL\,$\uparrow$
       & SR\,$\uparrow$ & SPL\,$\uparrow$ \\
\midrule
ZSON~\citep{zson}     & \ZSONHMSR & \ZSONHMSPL & \ZSONMPSR & \ZSONMPSPL \\
ESC~\citep{esc}       & \ESCHMSR & \ESCHMSPL & \ESCMPSR & \ESCMPSPL \\
L3MVN~\citep{l3mvn}   & \LThreeMVNHMSR & \LThreeMVNHMSPL & \LThreeMVNMPSR & \LThreeMVNMPSPL \\
VLFM~\citep{vlfm}     & \VLFMHMSR & \VLFMHMSPL & \VLFMMPSR & \VLFMMPSPL \\
SG-Nav~\citep{sgnav}  & \SGNavHMSR & \SGNavHMSPL & \SGNavMPSR & \SGNavMPSPL \\
ASCENT~\citep{ascent} & \ASCENTHMSR & \ASCENTHMSPL & \ASCENTMPSR & \ASCENTMPSPL \\
\midrule
\textbf{Ours (Full)} & \textbf{\OursHMSR} & \textbf{\OursHMSPL}
              & \textbf{\OursMPSR} & \textbf{\OursMPSPL} \\
\bottomrule
\end{tabular}
\caption{Ours leads every zero-shot baseline on both benchmarks and on both
metrics, so the added successes do not come from longer paths.}
\label{tab:main}
\end{table}

\paragraph{Benchmarks}
We evaluate on the ObjectNav \texttt{val} splits of HM3D v1~\citep{hm3d} and
MP3D~\citep{mp3d} in the Habitat simulator~\citep{habitat}. Both are built
from photorealistic scans of real buildings, so the robot moves through the
clutter and sight lines of houses that exist. HM3D v1 \texttt{val} contains
\NValHM{} episodes across
\NScenesHM{} buildings and \NCatsHM{} goal categories (chair, bed, plant,
toilet, sofa, TV monitor). MP3D \texttt{val} contains \NValMP{} episodes
across \NScenesMP{} buildings and \NCatsMP{} categories, among them small and
easily confused targets such as cushion, towel, and chest of drawers; its
buildings are larger and more often span several floors.

\paragraph{Metrics}
Success Rate (SR) is the fraction of episodes in which the robot calls stop
within \SuccDist{}\,m of an instance of the goal category, from a pose where
that instance can be viewed. Success weighted by Path Length
(SPL)~\citep{objectnav} scales each success by the ratio of the shortest
available path to the path actually walked, and scores zero on failure---SR
asks whether the object was found, SPL asks what the search cost. An episode
that reaches \MaxSteps{} steps ends in failure.

\subsection{Main Results}
\label{sec:main-results}

Ours (Full) leads every zero-shot baseline in Table~\ref{tab:main} on both
metrics and on both benchmarks: \OursHMSR\% SR and \OursHMSPL\% SPL on HM3D
v1 \texttt{val}, \OursMPSR\% SR and \OursMPSPL\% SPL on MP3D \texttt{val}.
Ours (Full) averages \OursHMCalls{} deliberative model calls and
\OursStepsPerEp{} environment steps per episode.

\subsection{Ablations}
\label{sec:ablations}

\begin{figure}[!t]
\centering
\includegraphics[width=\columnwidth]{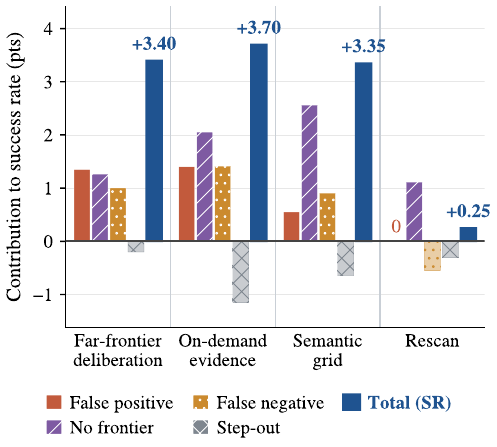}
\caption{Failure-class decomposition of each ablation gap on HM3D. Each group
is one component, measured by the Table~\ref{tab:ablation} row that takes it
away; within a group, the four coloured bars are that component's contribution
in SR points to each failure class and they sum to the blue \textbf{Total},
while bars below zero run the other way. Each component signs the spectrum
differently: deliberation spreads its \DeltaSRHM{} points over the three
decision-error classes, the semantic grid concentrates \GainMemNF{} of its
\GainMemTotal{} in no-frontier stops, and on-demand evidence
(\DeltaSROneShot{}) leads on both detection errors.}
\label{fig:failure}
\end{figure}

\begin{table}[t]

\centering
\small
\setlength{\tabcolsep}{0.8mm}
\begin{tabular}{@{}l ccccc@{}}
\toprule
Configuration & SR\,$\uparrow$ & SPL\,$\uparrow$
              & Calls & Img & Think\,(s) \\
\midrule
\textbf{Ours (Full)} & \textbf{\OursHMSR} & \textbf{\OursHMSPL}
        & \OursHMCalls & \OursHMImg & \OursHMThink \\
\midrule
Greedy Frontier  & \NeverHMSR & \NeverHMSPL
        & \NeverHMCalls & \NeverHMImg & \NeverHMThink \\
Always-Deliberate & \AlwaysHMSR & \AlwaysHMSPL
        & \AlwaysHMCalls & \AlwaysHMImg & \AlwaysHMThink \\
One-Shot          & \OneShotHMSR & \OneShotHMSPL
        & \OneShotHMCalls & \OneShotHMImg & \OneShotHMThink \\
w/o Semantic Grid & \NoMemHMSR & \NoMemHMSPL
        & \NoMemHMCalls & \NoMemHMImg & \NoMemHMThink \\
w/o Rescan        & \NoRescanHMSR & \NoRescanHMSPL
        & \NoRescanHMCalls & \NoRescanHMImg & \NoRescanHMThink \\
\bottomrule
\end{tabular}
\caption{Deliberation earns its gain through \emph{when} it fires and
\emph{what} it reads: firing only at structural impasses leads both on SR and
on SPL, while consulting the model at every frontier choice
(Always-Deliberate) scores below letting the value map choose (Greedy
Frontier). Each row changes one switch of the
full system: Greedy Frontier and Always-Deliberate change \emph{when}
deliberation fires, One-Shot \emph{how} the same image budget is used,
w/o~Semantic~Grid \emph{what} it may read, and w/o~Rescan \emph{when} to stop. Per
episode, \texttt{Calls} counts the deliberative controller's model calls and
\texttt{Img} how many of those calls carried images---the verification gate and
the floor decision run on budgets of their own---and \texttt{Think} is the
deliberation time.}
\label{tab:ablation}
\end{table}

\begin{figure}[t]
\centering
\includegraphics[width=\columnwidth]{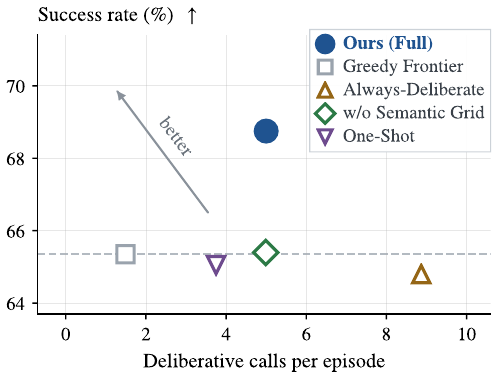}
\caption{The gain is not bought with deliberative calls. Always-Deliberate
makes the most calls per episode and scores lowest; One-Shot makes
fewer than Ours and lands on the dashed line---the Greedy Frontier success
rate---as does w/o~Semantic~Grid, which makes as many calls as Ours. Only the
full system, which fires at structural impasses \emph{and} reads
the anchored memory when it does, sits above that line. Up and to the left is
better, exact values in Table~\ref{tab:ablation}.}
\label{fig:pareto}
\end{figure}

Table~\ref{tab:ablation} reports five switch-level ablations and
Figure~\ref{fig:pareto} plots success against call count, every arm on the
same \NValHM{} episodes as Table~\ref{tab:main}. Greedy
Frontier---choosing among far frontiers by argmax instead of deliberating over
them---costs
\DeltaSRHM{} SR points, with a paired bootstrap 95\% confidence interval of
$[\DeltaSRHMCILo,\ \DeltaSRHMCIHi]$ and McNemar
$\chi^2{=}\McNemarChiHM$ ($p\,\McNemarPHM$).

Always-Deliberate holds every capability
fixed and removes only the trigger, waking the loop at every frontier choice
rather than at the far ones. It reaches \AlwaysHMSR\% SR---below Greedy
Frontier, which never deliberates over a frontier at all---costing
\DeltaSRAlways{} SR points against the full system
($[\DeltaSRAlwaysCILo,\ \DeltaSRAlwaysCIHi]$). It is given strictly more of
the model than the full system---more calls per episode, and more of them
image-bearing---and returns less success for it. Access to the model is
therefore not what produces the gain; the policy that decides when to spend it
is.
One-Shot keeps the trigger policy fixed but supplies
the images in one full-context query, isolating adaptive evidence
acquisition: it costs \DeltaSROneShot{} SR points
($[\DeltaSROneShotCILo,\ \DeltaSROneShotCIHi]$) while making \emph{fewer}
paid calls than the full system, every one of them image-bearing. What that
row gives up is buying evidence on demand, not access to imagery.
w/o~Semantic~Grid takes away the abstract tier alone---room types and confirmed
instances---and leaves the literal tier standing: keyframes are still written,
and the loop still recalls them. It spends the same budget as the full system
on both axes (\NoMemHMCalls{} against \OursHMCalls{} calls per episode,
\NoMemHMImg{} against \OursHMImg{} of them image-bearing) and still costs
\DeltaSRNoMem{} points
($[\DeltaSRNoMemCILo,\ \DeltaSRNoMemCIHi]$). Neither fewer calls nor less
imagery can explain that: one read surface is the only thing that changed.

\subsection{Analysis}
\label{sec:analysis}

\paragraph{Rescan attribution}
Rescan converts stored views into new navigation targets when the frontier is
exhausted: it fires on \RescanTrigRate\% of episodes, \RescanYieldRate\% of
which yield a new goal from a remembered view.

\paragraph{Verification gate}
The gate changes what the persistent map contains rather than merely flagging
doubtful entries: it vetoes \VerifyVetoRate\% of candidate detections before
fusion, so they never reach the object map.

\paragraph{Where the gain comes from}
Every episode ends in success or in exactly one of four failure classes, taken
from its recorded termination state rather than hand-labelled: two detection
errors (false positive, false negative) and two termination outcomes (no
frontier left, step-out). The per-class differences between an arm and the full
system therefore sum to their SR difference, and Figure~\ref{fig:failure} uses
that identity to attribute the gain.
Deliberation returns \GainDelibFP{}, \GainDelibNF{}, and \GainDelibFN{} points from false positives,
no-frontier stops, and false negatives---three classes at once, no single one
carrying the result. The semantic grid lands in one: \GainMemNF{}
of its \GainMemTotal{} points come from the premature stops it is built to
prevent. The two leave different signatures, and are complementary rather
than redundant. An episode kept alive by
suppressing a premature stop can only end in success or at the step cap, and
the split favours success: of the \GainDelibPos{} points deliberation redirects, \GainDelibTotal{}
arrive as success.

\section{Conclusion}
\label{sec:conclusion}
We presented a hierarchical fast--slow agent for zero-shot object-goal
navigation: an always-on reactive value-map controller writes a
coordinate-anchored memory, and an event-triggered, budget-bounded
ReAct agent reads it, reasoning from text first and retrieving imagery only
when text cannot decide. A pre-fusion gate keeps rejected detections out of the
map that authorizes stopping, and every deliberative failure returns control to
the reactive controller. It reaches \OursHMSR\% SR on HM3D v1
\texttt{val} and \OursMPSR\% on MP3D \texttt{val}, leading every
zero-shot baseline of Table~\ref{tab:main}; choosing among
far frontiers by argmax instead costs \DeltaSRHM{} SR points. What the robot has already
seen is the evidence worth reasoning over---and keeping that reasoning
bounded and fail-open is what makes it deployable.

\bibliography{refs}

\appendix
\setcounter{table}{0}
\setcounter{figure}{0}
\renewcommand{\thetable}{A\arabic{table}}
\renewcommand{\thefigure}{A\arabic{figure}}

\begin{figure*}[t]
\centering
\includegraphics[width=\textwidth]{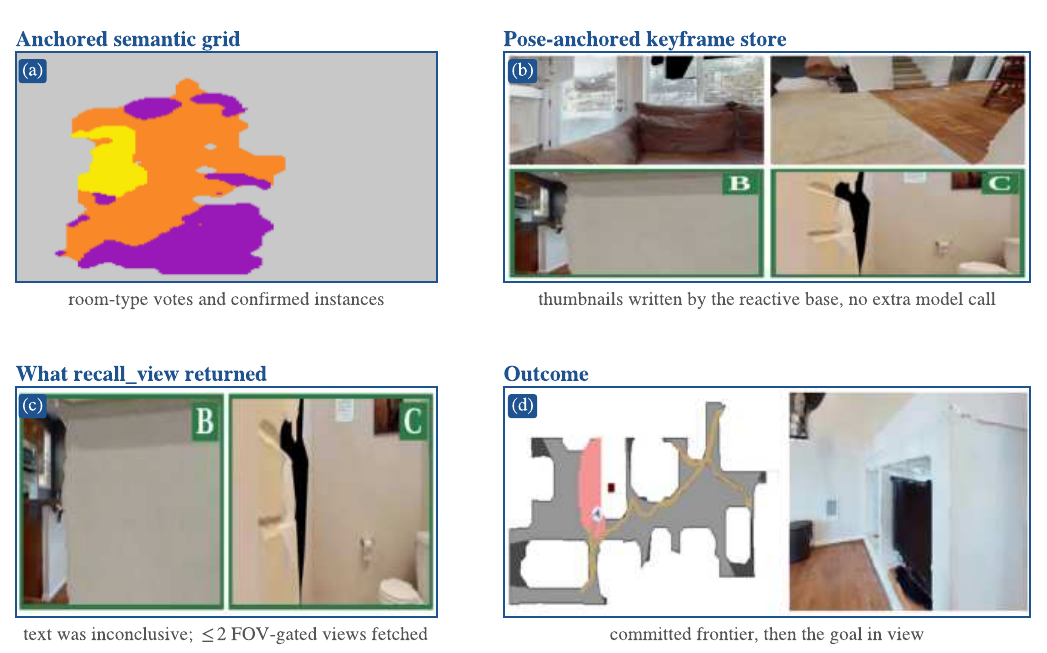}
\caption{One deliberation on one HM3D evaluation run (goal: TV,
success---the run of Figure~\ref{fig:memory}), at the distant-frontier
trigger.
\emph{(a)}~the coordinate-anchored semantic grid the reactive base had built
by the wake moment. Cells are tinted by their winning room-type vote---on
this floor hall (orange), kitchen (purple), and dining room (yellow)---over
light grey for space not yet observed. \emph{(b)}~the pose-anchored keyframe store at that
moment; the two views the loop went on to request are outlined and labelled
\textbf{B} and \textbf{C}. \emph{(c)}~the same two views as
\texttt{recall\_view} returned them, one per contested frontier, after text
alone could not separate the candidates.
\emph{(d)}~the committed frontier on the occupancy map, and the frame in
which the goal came into view.}
\label{fig:qual}
\end{figure*}

\section{Implementation of the Deliberative Loops}
\label{sec:appendix}

\begin{table}[!ht]

\centering
\small
\setlength{\tabcolsep}{3pt}
\begin{tabular}{@{}p{2.3cm}p{2.5cm}p{3.1cm}@{}}
\toprule
\textbf{Mode} & \textbf{Trigger} & \textbf{Per-invocation budget} \\
\midrule
\raggedright Recovery (tiered) & \raggedright contact accrual, or the secondary no-displacement test
(Section~\ref{sec:trigger}) & tier 0: replan, no call; model tier: $\le 3$ rounds, $\le 2$ images
in any one call \\
\addlinespace[2pt]
\raggedright Distant frontier & \raggedright nearest frontier beyond the distance gate (Section~\ref{sec:trigger}) & $\le 3$ rounds,
$\le 2$ images in any one call \\
\addlinespace[2pt]
\raggedright Inter-floor & \raggedright floor stalled, multiple floors known & single text call \\
\addlinespace[2pt]
\raggedright Rescan & \raggedright frontier set empty before reactive stop & single montage call,
capped per floor \\
\midrule
\multicolumn{3}{@{}p{7.9cm}@{}}{\emph{Per run:} $\le 20$ model calls, of
which $\le 10$ may be image-bearing; $\ge 15$ steps between re-triggers of
the same loop; rescan $\le 2$ per floor over a montage of $\le 8$ keyframes;
a per-step watchdog disables model calls for the rest of the run if any single
decision exceeds $180$\,s, and a circuit breaker disables them after
consecutive service failures.}\\
\bottomrule
\end{tabular}
\caption{Deliberation modes, triggering conditions, and the structure of
their budgets.}
\label{tab:triggers}
\end{table}

Each loop of Table~\ref{tab:triggers} is implemented as a declarative skill
specification---a Markdown file whose frontmatter names the loop's triggering
events, its model backend, its per-invocation budget, and the schema of the
typed command it must return---loaded at startup and dispatched by a single
runtime. Loops are therefore added, disabled, or re-budgeted without touching
the reactive base, and the trigger and budget columns of
Table~\ref{tab:triggers} are read directly from those files rather than
restated in code. A Python registry provides a fallback implementation for
any loop whose specification fails to load, which is one of the paths that
degrades to the reactive base.

\section{Computing Infrastructure}
\label{sec:infrastructure}

All experiments run on a workstation with a single NVIDIA GeForce RTX~3090
GPU (24\,GB), an Intel Core i7-14700KF CPU, and 32\,GB of RAM, under Ubuntu
20.04. The software stack is Python~3.9, PyTorch~2.1.0 with CUDA~11.8,
Habitat-Sim and Habitat-Lab~0.3.1, and ROS~Noetic. The perception models of
the reactive base---the image--text matching model, the open-vocabulary
detection pipeline, the segmentation and room-classification networks, and
the point-goal policy---run locally on the GPU; deliberation and the
verification gate invoke the external VLM service over its API, under the
budgets of Table~\ref{tab:triggers}.

\end{document}